\documentclass[letterpaper]{article} 
\usepackage{aaai2026}  
\usepackage{times}  
\usepackage{helvet}  
\usepackage{courier}  
\usepackage[hyphens]{url}  
\usepackage{graphicx} 
\usepackage{natbib}  
\usepackage{caption} 
\usepackage{algorithm}

\usepackage{algpseudocode}
\usepackage{amsmath}

\usepackage[utf8]{inputenc} 
\usepackage[T1]{fontenc}    
\usepackage{url}            
\usepackage{booktabs}       
\usepackage{amsfonts}       
\usepackage{nicefrac}       
\usepackage{microtype}      
\usepackage{xcolor}         
\usepackage{graphicx}
\usepackage{xspace}
\usepackage{tabularx}
\usepackage[table]{xcolor}
\usepackage{hyperref}

\usepackage{marvosym}
\usepackage{booktabs}

\definecolor{MyYellow}{RGB}{255, 230, 128}
\definecolor{DeepYellow}{RGB}{204, 153, 0}
\definecolor{SoftYellow}{RGB}{250, 240, 190}
\definecolor{light_blue}{rgb}{0.18,0.46,0.71}
\definecolor{gray_}{rgb}{0.52,0.8,0.97} 

\newcommand{\dataset}{ORS3D-60K\xspace}

\newcommand{\task}{ORS3D\xspace}
\newcommand{\model}{GRANT\xspace}
\usepackage{multirow}
\usepackage{pifont}
\definecolor{deepgreen}{rgb}{0.0, 0.5, 0.0}
\newcommand{\cmark}{{\textcolor{deepgreen}{\ding{51}}}}
\newcommand{\xmark}{\textcolor{red}{{\ding{55}}}}

\definecolor{cvprblue}{rgb}{0.21,0.49,0.74}

\definecolor{lightgray}{gray}{0.9}
\definecolor{linecolor}{rgb}{0.82, 0.94, 0.75}
\definecolor{mamba}{RGB}{153, 151, 239}
\definecolor{kaiming-green}{RGB}{57,181,74} 
\definecolor{pretty-blue}{RGB}{0, 113, 188}

\hypersetup{
    colorlinks=true,
    linkcolor=red,
    filecolor=magenta,      
    urlcolor=magenta,
    citecolor=kaiming-green
}

\usepackage{newfloat}
\usepackage{listings}
\DeclareCaptionStyle{ruled}{labelfont=normalfont,labelsep=colon,strut=off} 
\floatstyle{ruled}
\newfloat{listing}{tb}{lst}{}
\floatname{listing}{Listing}
\title{Cook and Clean Together: Teaching Embodied Agents for Parallel Task Execution}
\author {
    Dingkang Liang\textsuperscript{\rm 1$^{*}$},
    Cheng Zhang\textsuperscript{\rm 1\thanks{\footnotesize{Equal contribution. $\textsuperscript{\Letter}$ Corresponding author.}}},
    Xiaopeng Xu\textsuperscript{\rm 1},
    Jianzhong Ju\textsuperscript{\rm 2},
    Zhenbo Luo\textsuperscript{\rm 2},
    Xiang Bai\textsuperscript{\rm 1\Letter}
}
\affiliations {
    \textsuperscript{\rm 1}Huazhong University of Science and Technology\\
    \textsuperscript{\rm 2}MiLM Plus, Xiaomi Inc.\\
    \{dkliang, czhang2024, xbai\}@hust.edu.cn, \{jujianzhong, luozhenbo\}@xiaomi.com \\
    
}

\usepackage{bibentry}

\begin{document}

\maketitle

\begin{abstract}
Task scheduling is critical for embodied AI, enabling agents to follow natural language instructions and execute actions efficiently in 3D physical worlds. However, existing datasets often simplify task planning by ignoring operations research (OR) knowledge and 3D spatial grounding. In this work, we propose \textbf{O}perations \textbf{R}esearch knowledge-based 3D Grounded Task \textbf{S}cheduling (\task), a new task that requires the synergy of language understanding, 3D grounding, and efficiency optimization. Unlike prior settings, \task demands that agents minimize total completion time by leveraging parallelizable subtasks, e.g., \textit{cleaning the sink while the microwave operates}. To facilitate research on \task, we construct \dataset, a large-scale dataset comprising 60K composite tasks across 4K real-world scenes. Furthermore, we propose \model, an embodied multi-modal large language model equipped with a simple yet effective scheduling token mechanism to generate efficient task schedules and grounded actions. Extensive experiments on \dataset validate the effectiveness of \model across language understanding, 3D grounding, and scheduling efficiency.
\end{abstract}


\begin{links}
    \link{Code}{https://github.com/H-EmbodVis/GRANT}
\end{links}

\section{Introduction}
\label{sec:intro}

Task scheduling is fundamental for embodied agents to efficiently execute human-assigned tasks~\cite{embodied-survey,wang2023embodiedscan,leo,palme}. Achieving this requires the seamless integration of natural language understanding, efficiency optimization, and spatial perception within real-world 3D environments.

\begin{figure}[t]
  \centering
   \includegraphics[width=0.92\linewidth]{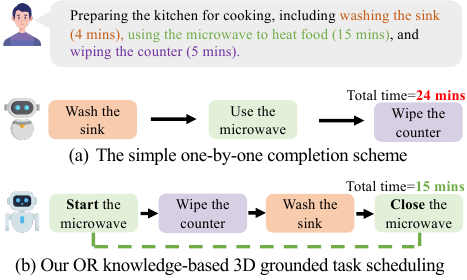}
   \vspace{-5pt}
   \caption{Comparison of different task completion schemes. An embodied agent is expected to use operations research knowledge to efficiently complete tasks through scheduling.}
   \label{fig:intro}
\end{figure}

Recently, several works~\cite{leo, grounded3dllm, sg3d} have made preliminary attempts on plan generation in 3D environments, allowing models to generate step-by-step plans from human instructions (Fig.~\ref{fig:intro}(a)). Nevertheless, these attempts are oversimplified and exhibit critical limitations that hinder their practical applications. First, they lack consideration of task properties and optimization of efficiency. Under their setting, a model only needs to generate plausible actions in terms of natural language.
In contrast, as shown in Fig.~\ref{fig:intro}(b), an embodied agent is assumed to have the capacity to efficiently complete the task by leveraging \textbf{O}perations \textbf{R}esearch (OR) knowledge. This includes identifying which subtasks can be executed concurrently with other subtasks and maximizing the use of waiting time to achieve optimal efficiency. 
Second, although their setting assumes an agent operating in 3D environments, it is often reduced to textual question answering, without explicitly grounding each step to the target object's location within the 3D scene.
This lack of spatial grounding severely hinders the utility of such plans for downstream embodied executions that require spatial location information (e.g., navigation).

\begin{figure*}[t]
  \centering
   \includegraphics[width=0.98\linewidth]{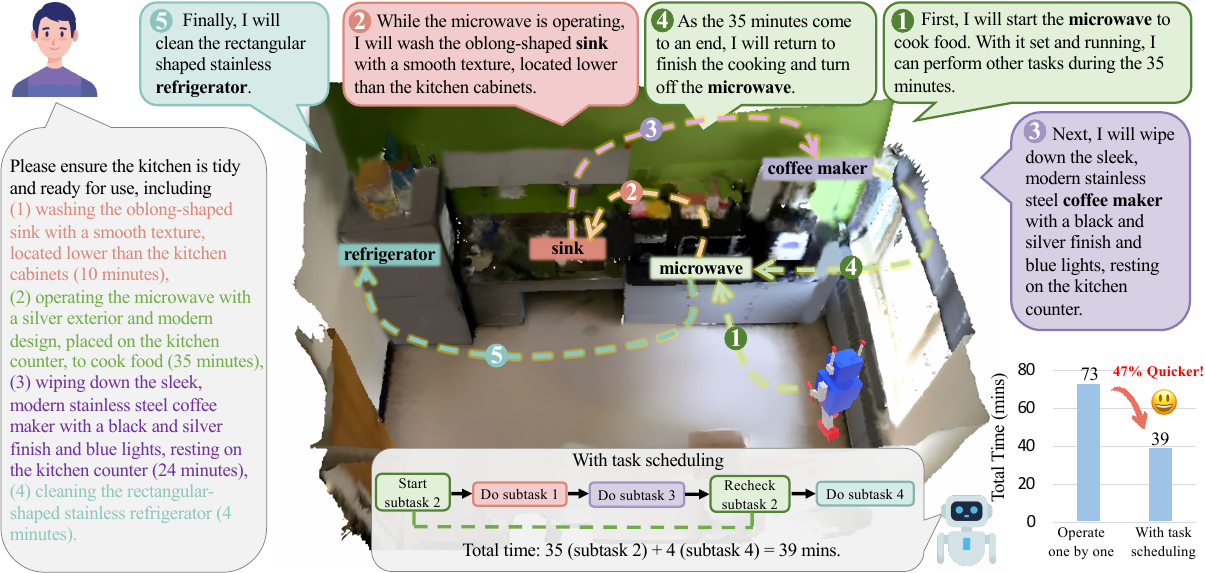}
    \vspace{-5pt}
   \caption{Illustration of the proposed Operations Research knowledge-based 3D Grounded Task Scheduling (ORS3D). When assigned a composite task by a human, the embodied agent needs to complete the subtasks efficiently by carefully scheduling using operations research knowledge and simultaneously locating the target objects in each step for navigation and manipulation.}
   \label{fig:task_illustration}
   \vspace{-10pt}
\end{figure*}

To address these limitations and extend the capability of embodied agents for efficient task scheduling, we propose a new and practical task named \textbf{O}perations \textbf{R}esearch knowledge-based 3D Grounded Task \textbf{S}cheduling (\textbf{ORS3D}). In this task, an embodied agent must generate efficient schedules by leveraging OR knowledge and locate the 3D positions of target objects in each action step to complete assigned tasks. 
As demonstrated in Fig.~\ref{fig:task_illustration}, when we assign a composite task to an embodied agent, we hope it can efficiently complete it by utilizing the waiting periods of subtasks that can be performed concurrently. For example, "Using the microwave" allows the agent to perform other subtasks during its waiting period. 
To achieve maximum efficiency, the embodied agent must leverage these subtask properties and incorporate OR knowledge to generate an optimal task schedule. Meanwhile, to execute each step in the real world, the agent must accurately localize the target objects within the 3D scene. Therefore, ORS3D poses significant challenges to existing 3D agents~\cite{leo, chatscene, 3dllava,grounded-task,grounded3dllm} in two essential aspects: 
\textbf{1)} It requires OR knowledge to identify subtasks that can be performed concurrently and make efficient task schedules. 
\textbf{2)} It entangles language and spatial understanding (i.e., an embodied agent is required to simultaneously generate actions and locate the target objects in the 3D scene).

To facilitate research on this new task, we construct the \dataset dataset consisting of 60,825 composite tasks across 4,376 real-world indoor scenes. As shown in Tab.~\ref{tab:data_compare}, compared to existing 3D understanding and task planning-related datasets~\cite{TaPA, grounded3dllm, sg3d, leo, scanreason}, \dataset is the first to incorporate OR knowledge. 
It also has the largest number of tasks and presents the most significant challenge by requiring models to generate lengthy textual solutions and provide 3D grounding for target objects.
To assess the capability of existing methods in addressing this task, we evaluate several baselines~\cite{leo,pq3d,3d-vista,grounded3dllm} from language understanding, efficiency optimization, and spatial perception, where the results show they struggle with this challenging task.

To tackle the \task problem, we further propose a \underline{\textbf{gr}}ounded t\underline{\textbf{a}}sk scheduling age\underline{\textbf{nt}} named \textbf{\model}, which is empowered by a Multi-modal Large Language Model (MLLM) and equipped with a simple yet effective Scheduling Token Mechanism (STM) to generate efficient task schedules.
Specifically, we introduce a learnable scheduling token that links to an external optimization solver to generate task schedules based on task property constraints provided by the MLLM. The solver employs a dynamic programming algorithm to arrange subtasks within the available time intervals of those that can be performed concurrently, producing an optimal execution schedule.
During inference, \model first predicts subtask properties as constraints, then uses the scheduling token to invoke the solver and generate the optimal schedule, which is subsequently injected back into the model to guide the generation of step-wise action descriptions and target object groundings.
Compared to the baseline method~\cite{grounded3dllm}, our approach yields a significant 30.53\% improvement in task completion time efficiency, along with notable gains of 1.38\% in grounding accuracy and 10.46\% in overall performance. 
As an initial attempt, our method paves the way for further exploration in \task. 

\begin{table*}[t]
\centering
\footnotesize
\setlength{\tabcolsep}{0.7mm}
\begin{tabular}{l ccc ccc ccc}
\toprule
Dataset & Reference & \#Scenes & \#Task  & Avg. length & Text output  & 3D Grounding & Planning & OR knowledge \\
\midrule
TaPA~\cite{TaPA}  & arXiv 23  & 80 & 15,418 & 69 & \cmark  & \xmark  & \cmark & \xmark \\
Embodied planning~\cite{grounded3dllm} & arXiv 24  & 1,319 & 4,357 & 37 & \cmark  & \cmark  & \cmark & \xmark \\
SG3D~\cite{sg3d}   & arXiv 24   & 4,895 & 22,346  & 71 & \xmark   & \cmark   & \cmark & \xmark\\
ScanReason~\cite{scanreason} & ECCV 24  &  1,456 & 12,929 & 29 & \xmark  & \cmark & \xmark & \xmark\\
LEO (Task planning)~\cite{leo}      & ICML 24    & 478  & 13,848  & 98  & \cmark  & \xmark & \cmark & \xmark\\
Intent3D~\cite{intent3d} & ICLR 25 &  1,042 & 44,990 & 9 & \xmark  & \cmark  & \xmark & \xmark\\
\midrule
\rowcolor{cyan!15}
\textbf{\dataset (ours)} & -  & 4,376 & \textbf{60,825}  & \textbf{311} & \cmark  & \cmark & \cmark  & \cmark \\
\bottomrule
\end{tabular}
\vspace{-5pt}
\caption{Comparison with related datasets. "Avg. length" denotes the average word length of each data item. Our dataset is the only one that introduces Operations Research (OR) knowledge for task scheduling.}
\label{tab:data_compare}
\end{table*}

In summary, our contributions are as follows: 
1) We introduce operations research knowledge-based 3D grounded task scheduling, a new and practical task that meets the common requirement of embodied agents to efficiently complete tasks in the physical world.
2) To support this new task, we construct a large-scale dataset, \dataset. To the best of our knowledge, we are the first to incorporate operations research knowledge for task scheduling in 3D scenarios.
3) We propose \model, an embodied MLLM with a simple yet effective scheduling token mechanism, integrating task scheduling with multimodal understanding to generate efficient, grounded task execution schedules.

\section{Related Works}
\label{sec:related}

\subsection{Task Planning}

Task Planning~\cite{saycan, LoTa-Bench, sg3d, EgoPlan-Bench} is crucial, as it enables embodied agents to execute human instructions efficiently. Wu et al.~\cite{TaPA} propose TaPA, a vision-language task planning agent that generates executable textual action steps for robot navigation and manipulation using multi-view images of the 3D scene. Huang et al.~\cite{leo} construct a task planning dataset that requires embodied agents to generate step-wise plans from instructions.
SG3D~\cite{sg3d} proposes task-oriented sequential grounding in 3D scenes, where an agent is required to locate each target object in a given sequence of actions.
In contrast to previous works, we focus on more complex scheduling scenarios and the integration of multi-modal information processing.

\subsection{3D Scene Understanding} 
3D scene understanding is the foundation of embodied AI, enabling it to act in real-world scenes. 3D scene understanding includes depth estimation~\cite{xu2023accurate,xu2025igev++}, 3D object detection~\cite{unidet3d,liang2025sood++,zhou2025hermes}, segmentation~\cite{takmaz2023openmask3d,uniseg3d,liang2024pointmamba,liang2025parameter}, and grounding~\cite{scanrefer, reason3d, more3d}. Mask3D~\cite{mask3d} is often used as an off-the-shelf object proposal extractor for downstream tasks or as a 3D scene encoder, as the flexible learned instance queries can be easily assembled to Transformer-based LLMs. 
OneFormer3D~\cite {oneformer3d} is an end-to-end method that performs instance and semantic segmentation consistently, utilizing a group of learnable instance queries.

\subsection{3D Multi-modal Large Language Models }

3D MLLMs~\cite{grounded3dllm,chat3d, chatscene,robin3d,fu2025orion,integratingchainofthought,pq3d,llava3d,3dllm} narrow the gap between spatial understanding and natural language processing. Several methods~\cite{leo,3d-vista,robin3d,pq3d} utilize point cloud object proposals from off-the-shelf 3D object detectors to extract 3D scene information.
Another line of research~\cite{3dllm,llava3d} leverages pretrained 2D encoders to reconstruct 3D information for the LLMs.
In contrast, other approaches like Grounded 3D LLM~\cite{grounded3dllm} and 3D-LLaVA~\cite{3dllava} directly process scene point clouds using 3D scene encoders that are jointly trained with LLMs.
However, although existing 3D MLLMs excel at scene understanding, they still lack the ability to leverage OR knowledge for efficient task scheduling and completion.

\section{The \dataset Dataset}
\label{sec:dataset}
In this section, we introduce the definition of \textbf{O}perations \textbf{R}esearch knowledge-based 3D Grounded Task \textbf{S}cheduling (\task) and provide details of the proposed \dataset dataset.

\begin{figure}[ht]
  \centering
   \includegraphics[width=\linewidth]{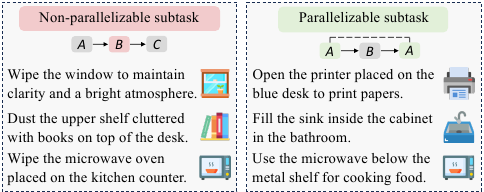}
   \vspace{-15pt}
   \caption{Non-parallelizable subtask \& parallelizable subtask.}
   \label{fig:task}
\end{figure}

\begin{figure*}[ht]
  \centering
   \includegraphics[width=0.98\linewidth]{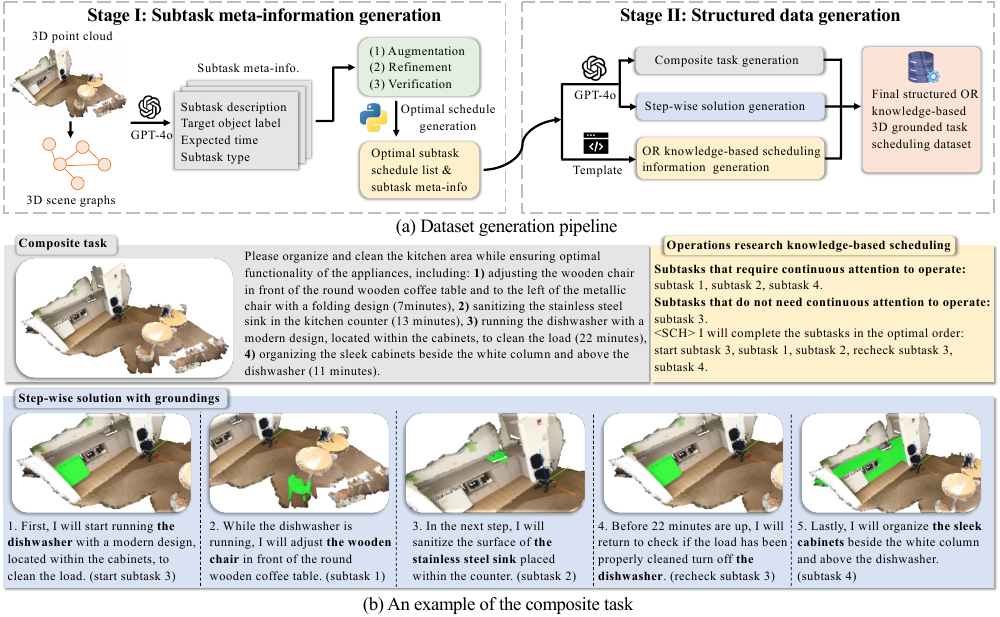}
   \vspace{-5pt}
   \caption{(a) The \dataset dataset generation pipeline, which first generates subtask meta-information from 3D scene graphs, then uses this information to generate the structured dataset. (b) A composite task example from \dataset dataset. The green color mask indicates the ground-truth target object in the corresponding step.}
   \label{fig:data_engine}
\end{figure*}

\subsection{Design Principles}
\label{sec:principles}

Understanding human instructions, making efficient schedules to complete human-assigned tasks, and interacting with objects are common and frequent requirements in real-world applications for embodied agents.

As illustrated in Fig.~\ref{fig:task}, tasks assigned to embodied agents can be categorized into two types from an OR perspective: 
1) \textbf{Non-parallelizable subtask} requires continuous attention of the agent to manipulate the target object, such as wiping the table or dusting the shelf. 2) \textbf{Parallelizable subtask} only requires the agent to initiate and recheck the target object upon completion, without continuous attention and manipulation, such as using the microwave to heat food or filling the water sink. The agent needs to exploit the time intervals of parallelizable subtasks to achieve an efficiency objective.


\subsection{Problem Formulation}
\label{sec:Problem}
The goal of OR knowledge-based 3D grounded task scheduling is to generate an efficient schedule and accurately locate the target object at each step to complete a composite task.

Specifically, suppose that an embodied agent in a 3D scene is assigned a composite task consisting of \(n\) subtasks, denoted as \(\mathcal{C}=\{\tau_i \}_{i=1}^n\). Each subtask $\tau_i$ is an operation involving a target object with an expected time, described by a natural language instruction. 
To achieve efficient task scheduling, the agent needs to generate a time-efficient schedule \(\mathcal{A}=\{a_i|(\tau_i, l_i)\}_{i=1}^s\) consisting of \(s\) steps to accomplish the composite task. Each step includes a textual action description \(a_i\) for subtask $\tau_i$ and the 3D location \(l_i\) (e.g., 3D bounding box or point mask) of the target object.

\subsection{Dataset Construction}
\label{sec:Construction}
The dataset construction pipeline is illustrated in Fig.~\ref{fig:data_engine}(a).
In Stage I, we use 3D point clouds from five real-world datasets: ScanNet~\cite{scannet200}, HM3D~\cite{hm3d}, ARKitScenes~\cite{baruch2021arkitscenes}, 3RScan~\cite{wald2019rio}, and MultiScan~\cite{mao2022multiscan}. They are converted into textual 3D scene graphs~\cite{jia2024sceneverse} for subtask meta-information generation via GPT-4o. We refine the outputs for correctness and completeness, and perturb subtask expected times by $\pm 10\%$ to generate diverse optimal schedules.
In Stage II, we compute the optimal task schedule using an optimization solver, then convert it into step-wise natural language instructions with phrase-level object grounding via GPT-4o. We also generate OR knowledge-based scheduling explanations using templates.

Fig.~\ref{fig:data_engine}(b) presents a data example from the \dataset dataset. The composite task comprises a list of subtasks that the embodied agent must complete. The solution consists of step-by-step actions with target object locations. At each step, the model is required to simultaneously produce an action description of the operation on a subtask and locate the target object in the 3D scene.

\subsection{Dataset Characteristics}
\label{sec:Characteristics}

\begin{figure}[t]
  \centering
   \includegraphics[width=0.99\linewidth]{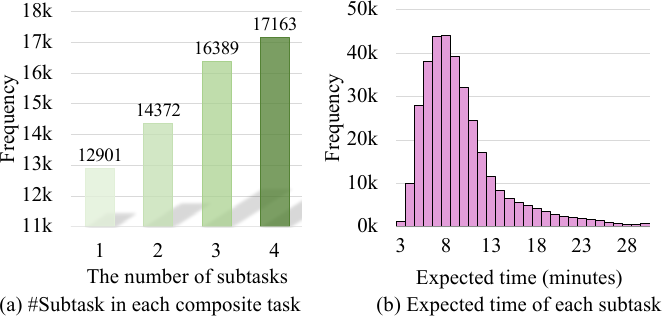}
      \vspace{-5pt}
   \caption{ Distributions of (a) subtask number in each composite task, and (b) the expected time of each subtask.}
   \label{fig:data_haracteristics}
\end{figure}

\begin{figure*}[t!]
  \centering
   \includegraphics[width=0.98\linewidth]{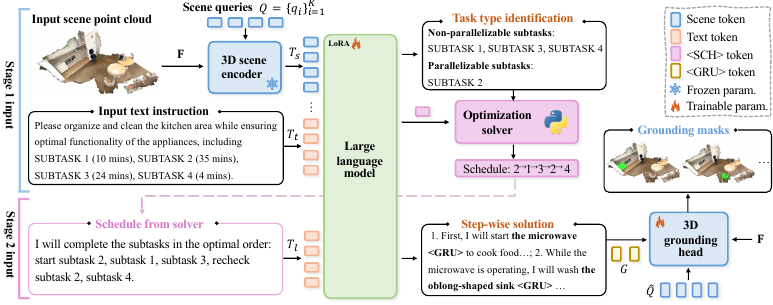}
   \vspace{-5pt}
   \caption{Overview of \model. The scene point cloud is processed by a 3D scene encoder into scene tokens. \model first infers task properties (stage 1), then uses a scheduling token to generate an optimal schedule (stage 2). The grounding tokens are fed to the 3D grounding head to generate object masks. The input task description is simplified for brevity.}
   \label{fig:main}
\end{figure*}

The \dataset dataset exhibits several distinctive characteristics that make it stand out from existing datasets.

First, our dataset is closely aligned with real-world task-completion scenarios, extending beyond existing 3D visual grounding and question-answering datasets~\cite{scanrefer, intent3d, scanreason}. It is characterized by the inclusion of OR knowledge, which is not considered in existing related 3D understanding datasets. 

Second, as shown in Tab.~\ref{tab:data_compare}, our dataset has an exceptionally high average text length of 311 words, which poses a significant challenge for the language processing capabilities of embodied agents. Our dataset contains 60,825 composite tasks across 4,376 scenes, representing the largest scale among existing related datasets. 
Besides, as shown in Fig.~\ref{fig:data_haracteristics}(a), our dataset covers different levels of difficulty, reflected by the varying number of subtasks (4 to 7) in each composite task. The expected time for each subtask (Fig.~\ref{fig:data_haracteristics}(b)) follows a long-tail distribution, reflecting real-world variability.
Furthermore, in our \task setting, 3D grounding is entangled within the text, requiring the model not only to understand what actions to take, but also to accurately identify where each action should occur in the 3D scene. These characteristics make our dataset large-scale, diverse in task complexity, and highly representative of real-world scenarios.

\section{Method}
\label{sec:method}
The \task task requires integration of language understanding, efficiency optimization, and spatial perception in 3D environments. Therefore, we propose a \underline{\textbf{gr}}ounded t\underline{\textbf{a}}sk scheduling age\underline{\textbf{nt}}, termed \textbf{GRANT}, where the overall architecture is illustrated in Fig.~\ref{fig:main}. 
Specifically, \model consists of four components: 1) A 3D scene encoder converting point clouds into scene tokens. 2) An LLM processing multimodal inputs for task understanding. 3) A scheduling token mechanism (STM) connecting the LLM to an optimization solver for efficient scheduling. 4) A grounding head generating point masks for object localization. The scene encoder and grounding head integrate with the LLM through specialized tokens, enabling end-to-end training for simultaneous schedule generation and spatial grounding.

\subsection{Multi-modal Input Processing}

\model takes both the scene point cloud and the textual description as input, which are first tokenized and subsequently fed into the LLM for unified understanding.

\noindent\textbf{Point cloud tokenization.}
For a scene point cloud $\mathcal{P} \in \mathbb{R}^{N  \times 6}$, where each point contains 6-dimensional information $[x, y, z, r, g, b]$ and $N$ is the number of points, a sparse convolutional network is employed to extract point-wise features $\mathbf{F} \in \mathbb{R}^{N \times d}$, where $d$ is the feature dimension. We then use a pre-trained 3D scene encoder to further encode the point cloud features into scene tokens.
The 3D scene encoder $\mathcal{E}$ employs a fixed set of $K$ learnable scene queries $Q = \{\boldsymbol{q}_i\}_{i=1}^K$, which interact with point cloud features via cross-attention to produce processed scene queries containing rich semantic information. This process can be formulated as:
\begin{equation}
\hat{Q} = \mathcal{E}(Q, \mathbf{F}),
\end{equation}
\noindent where $\hat{Q} = \{\hat{\boldsymbol{q}}_i\}_{i=1}^K$ is the processed scene queries.
To align with the token embedding dimension $D$ of the LLM, the processed scene queries are projected into scene tokens \( T_s = \{\boldsymbol{s}_i\}_{i=1}^K \) via a simple linear layer, where each \( \boldsymbol{s}_i \in \mathbb{R}^{1 \times D} \) represents a scene token.

\noindent\textbf{Text tokenization.} We employ a text tokenizer to convert the input composite task description into a sequence of text tokens $T_t = \{\boldsymbol{x}_i\}_{i=1}^L$, where each $\boldsymbol{x}_i \in \mathbb{R}^{1 \times D}$ and $L$ denotes the length of the input text.

\noindent\textbf{LLM processing.}
The LLM plays a central role in our model by handling multi-modal inputs and understanding both point clouds and human instructions. It further identifies subtask types, solves complex task scheduling problems, generates descriptive action steps, and provides 3D positions of target objects in a unified manner.
As shown in Fig.~\ref{fig:main}, the scene tokens are prepended to the text tokens and fed to the LLM, which generates output tokens in an auto-regressive manner. The output tokens include specially designed tokens for task scheduling and 3D grounding, which will be elaborated in the following sections.

\begin{algorithm}
\footnotesize
\caption{Optimization-based Scheduling Solver (single parallelizable subtask)}
\label{alg:1}
\begin{algorithmic}[1]
\State \textbf{Input:} $\mathcal{I} = \{ (\tau_i, c_i, t_i) \}_{i=1}^n$
\State \textbf{Output:} Schedule $S^*$
\State Split subtasks into one parallelizable subtask $\tau_{\text{P}}$ (if any) and non-parallelizable set $S_{\overline{\text{P}}}$
\If{no parallelizable subtask}
    \State $S^* \gets S_{\overline{\text{P}}}$ \Comment{Purely sequential}
\Else
    \State $T_{\text{P}} \gets$ duration of $\tau_{\text{P}}$
    \State Select $S_{\text{in}} \subseteq S_{\overline{\text{P}}}$ s.t. $\sum_{\tau_i \in S_{\text{in}}} t_i \le T_{\text{P}}$ and the sum is maximized
    \State $S_{\text{out}} \gets S_{\overline{\text{P}}} \setminus S_{\text{in}}$
    \State $S^* \gets S_{\text{out}} \;+\; [\tau_{\text{P}}] \;+\; S_{\text{in}} \;+\; [\tau_{\text{P}}]$
\EndIf
\State \Return $S^*$
\end{algorithmic}
\end{algorithm}

\subsection{Scheduling Token Mechanism}

LLMs exhibit strong capabilities in natural language generation but are generally less effective at solving complex mathematical problems. To address this limitation, we introduce a special $\texttt{<SCH>}$ token that connects to an external solver to obtain an optimal scheduling list. This list is then utilized to guide the LLM for step-wise action generation.
Specifically, for a composite task description, the LLM first identifies the parallelizable and non-parallelizable subtasks (defined in Sec.~\ref{sec:principles}). 
It then constructs the subtask type information as $\mathcal{I} = \{  (\tau_i, c_i, t_i)\}_{i=1}^n$, where $c_i \in \{\text{P},\overline{\text{P}}\}$ denotes the subtask type (P: parallelizable, $\overline{\text{P}}$: non-parallelizable) and $t_i$ is the expected time.

The information $\mathcal{I}$ is passed to an external optimization solver via the $\texttt{<SCH>}$ token. As defined in Alg.~\ref{alg:1}, the solver minimizes the total execution time given the subtask types and their expected times. This is formulated as a 0–1 knapsack problem, where the waiting interval of a parallelizable subtask plays the role of the capacity and the durations of non-parallelizable subtasks serve as item weights and values, so that the solver maximizes the utilization of the waiting time of parallelizable subtasks while minimizing the overall completion time.

A simple dynamic programming algorithm is employed to solve this problem. The solver finally returns the optimal schedule of subtask IDs.
This scheduling process can be formulated as:
\begin{equation}
    S^* = \text{Solver}(\mathcal{I}),
\end{equation}
where \( S^* \) is the optimal subtask completion schedule, represented as a list of subtask IDs.
Then, $S^*$ is converted into natural language using predefined templates, then tokenized into $T_l$ by the text tokenizer and concatenated with the preceding tokens to guide the LLM in generating step-wise action descriptions.

\subsection{3D Grounding Head}

Besides generating action descriptions, the model also needs to simultaneously locate the corresponding target object in order to complete the task in the physical world.
To achieve this, we use a special $\texttt{<GRU>}$ token to indicate the target object for grounding in the output of LLM. To align with the dimension of the processed scene queries, all output $\texttt{<GRU>}$ tokens are passed through a simple MLP head into $G=\{\boldsymbol{g}_j\}_{j=1}^s$, where each $\boldsymbol{g}_j \in \mathbb{R}^{1 \times d}$. 

The target scene query is selected through max cosine similarity. Specifically, we compute the cosine similarity between $\boldsymbol{g}_j$ and each $ \hat{\boldsymbol{q}}_i $. The scene query with the highest probability is selected as the best match one, denoted as $\boldsymbol{q}^*\in \mathbb{R}^d $. This process can be formulated as:

\begin{equation}
\boldsymbol{q}^* = \arg\max_{\hat{\boldsymbol{q}}_i \in \hat{Q}} \;\left( \frac{\boldsymbol{g}_j \cdot \hat{\boldsymbol{q}}_i}{\|\boldsymbol{g}_j\| \cdot \|\hat{\boldsymbol{q}}_i\|} \right),
\end{equation}

The grounding mask is generated by the matched scene query with point cloud features. The mask corresponding to the scene query $\boldsymbol{q}^*$ is computed by taking the dot product between $\boldsymbol{q}^*$ and the point cloud features, followed by a sigmoid activation to obtain a point mask, which is expressed as:
\begin{equation}
\boldsymbol{m} = \sigma\left( \mathbf{F} \cdot \boldsymbol{q}^* \right),
\end{equation}
where \( \boldsymbol{m} \in \mathbb{R}^N \) is the predicted point mask of target object.

\noindent\textbf{Training objectives.} 
For language modeling, we use next-token prediction with cross-entropy loss. For grounding, we align grounding tokens and scene queries via a similarity matrix and supervise it using a binary correspondence matrix with sigmoid focal loss.

\begin{table*}[t]
\centering
\scriptsize
\setlength{\tabcolsep}{2.2mm}
\begin{tabular}{lcccccccc}
\toprule 
 \multirow{2.3}{*}{Method} & \multirow{2.3}{*}{Venue} & \multirow{2.3}{*}{3D Obj. Det.} & \multirow{2.3}{*}{LLM} & \multicolumn{2}{c}{Language} & \multicolumn{1}{c}{Scheduling}   & \multicolumn{1}{c}{3D Grounding}   & \multirow{2.3}{*}{Overall $\uparrow$}  \\ 
  \cmidrule(lr){5-6}\cmidrule(lr){7-7}\cmidrule(lr){8-8}
 & &  & &METEOR $\uparrow$ & ROUGE $\uparrow$ & TE $\uparrow$ & Accuracy $\uparrow$  & \\ 
\midrule
\rowcolor{gray!20}
\multicolumn{9}{l}{\textit{Commercial LLM/MLLMs (only text input)}}  \\
\midrule
Gemini    & -  & -& Gemini-2.0-flash   & 41.67  &  58.48 & 24.75  & \multicolumn{1}{c}{\multirow{3}{*}{Unsupported}} &  31.22\\
DeepSeek-R1~\cite{deepseekr1}   & -  & -& DeepSeek-V3  & 32.40 & 41.50  & 72.63  &   & 36.63\\ 
GPT-4o    & - & - & GPT-4o   & 49.16 & 62.19 & 45.27   & &   39.15\\
\midrule
\rowcolor{gray!20}
\multicolumn{9}{l}{\textit{Object-level methods (with detected object proposals*)}} \\
\midrule
3D-VisTA~\cite{3d-vista} & ICCV 23 &  Mask3D  & - & \multicolumn{3}{c}{\multirow{2}{*}{Unsupported}} 
 & \textcolor{gray}{54.90}$^\ddag$      & 13.73\\ 
PQ3D~\cite{pq3d} & ECCV 24  & Mask3D  & - & & & & \textcolor{gray}{56.12}$^\ddag$      &14.03\\ 
LEO$^\dag$~\cite{leo}   & ICML 24 & Mask3D & Vicuna-1B  & 46.61 & 60.32 & 45.63 & Unsupported & 38.14\\
\midrule
\rowcolor{gray!20}
\multicolumn{9}{l}{\textit{\textbf{Scene-level methods}}} \\
\midrule
Grounded 3D LLM~\cite{grounded3dllm}  &	arXiv 24  &  -& Vicuna-1B   & 41.96 & 53.71  & 42.46 & 34.00  & 43.03\\ 
\model (\textbf{ours}) & -  &  - & Vicuna-1B   & \textbf{42.82 }& \textbf{62.78} & \textbf{72.99}  & \textbf{35.38} & \textbf{53.49}\\
\bottomrule
\end{tabular}
\vspace{-5pt}
\caption{Experiment results on ~\dataset test set. $^\dag$ We adapt LEO by replacing its LLM with Vicuna-1B for a fair comparison. * indicates that these methods require object point clouds from an external 3D detector like Mask3D~\cite{mask3d}. $^\ddag$ Results are produced by directly providing step-wise schedules as input. Overall is the average of METEOR, ROUGE, TE, and Grounding Accuracy (treating unsupported metrics as 0). }
\label{tab:main}
\end{table*}

\begin{table*}[t]
\centering
\scriptsize
\setlength{\tabcolsep}{1.5mm}
\begin{minipage}{0.48\textwidth}
\centering
\begin{tabular}{lcccc}
\toprule
\multirow{2.3}{*}{Method} & \multirow{2.3}{*}{3D Obj. Det.} & \multicolumn{2}{c}{Detection} & \multicolumn{1}{c}{Segmentation} \\ 
\cmidrule(lr){3-4}\cmidrule(lr){5-5}
 & & AP\hspace{1pt}@0.25 & AP\hspace{1pt}@0.50 & mIoU  \\ 
\midrule
3D-VisTA*    & Mask3D & 54.90 & 41.88 & 43.29  \\ 
PQ3D* & Mask3D & 56.12 & 44.01 & 46.37  \\ 
\midrule
Grounded 3D LLM     & -    & 34.00  & 23.93 & 25.56  \\ 
\model (\textbf{ours})  & -  & \textbf{35.38}  & \textbf{24.79} & \textbf{26.71} \\ 
\bottomrule
\end{tabular}
\vspace{-5pt}
\caption*{(a) 3D grounding performance comparison}
\label{tab:grounding}
\end{minipage}
\hfill
\begin{minipage}{0.48\textwidth}
\centering
\setlength{\tabcolsep}{0.6mm}
\renewcommand{\arraystretch}{1.3}
\begin{tabular}{lcccccccc}
\toprule
 \multirow{2.3}{*}{Method} & \multirow{2.3}{*}{Acc.}  & \multicolumn{3}{c}{Para. subtask} & \multicolumn{3}{c}{Non-para. subtask}   & \multicolumn{1}{c}{Scheduling}    \\ 
  \cmidrule(lr){3-5}\cmidrule(lr){6-8} \cmidrule(lr){9-9}
&  &Prec.   & Recall   & F1   &Prec.   & Recall   & F1   & TE    \\ 
\midrule
Grounded 3D LLM     & 77.14  & 73.80 & 50.15 & 59.72 & 95.17 & 82.66 & 88.48 & 42.46\\ 
LEO & 79.73 & \textbf{78.19} & 41.57 & 54.28 &95.90 & 87.37  & 91.43 & 45.63 \\ 
\model (\textbf{ours})  & \textbf{84.65}  & 73.82  &\textbf{ 54.70} & \textbf{62.84} & \textbf{95.94} &  \textbf{90.67} & \textbf{93.23} & \textbf{72.99}\\ 
\bottomrule
\end{tabular}
\vspace{-5pt}
\caption*{(b) Subtask type recognition and scheduling}
\label{tab:type}
\end{minipage}

\vspace{-5pt}
\caption{(a) Comprehensive 3D grounding performance. * indicates that these methods require object point clouds from an external 3D detector like Mask3D~\cite{mask3d}. (b) Impact of subtask type recognition on scheduling efficiency.}
\label{tab:grounding_type}
\end{table*}

\section{Experiments}
\label{sec:Experiments}

\subsection{Implementation Details} 
The 3D scene encoder of \model is initialized with a pretrained CLASP~\cite{grounded3dllm}, freezing all weights except the projection layer used for alignment. We use Tiny-Vicuna-1B~\cite{vicuna} as the LLM and fine-tune it using LoRA~\cite{hu2021lora}. 
We use the AdamW optimizer with a cosine learning rate scheduler (initialized as \(8 \times 10^{-4}\)) and a weight decay of 0.1. Models are trained for 10 epochs on the \dataset training set with a batch size of 1. All experiments are conducted on 8$\times$ RTX 4090 GPUs.

\subsection{Evaluation Metrics}

The model performance is evaluated across three aspects that align with the challenges of the \task task. For output language quality, we use NLP metrics (METEOR \& ROUGE). For 3D grounding accuracy on the target object at each step, we adopt the AP@25\% detection metric. Considering that the core aspect of the \task task is \textit{scheduling}, we introduce the \textbf{Time Efficiency (TE)} metric to measure how well the model utilizes the time intervals of parallelizable subtasks. Rather than using raw completion time, TE normalizes the efficiency of each schedule between a naive sequential baseline and the optimal schedule. Formally, the TE of a predicted task schedule is calculated as:
\begin{equation}
    \mathrm{TE} = \frac{ \mathcal{T}_{\text{worst}} - \mathcal{T}_{\text{pred}} }{ \mathcal{T}_{\text{worst}} - \mathcal{T}_{\text{opt}} } \times 100\%,
\end{equation}
where $\mathcal{T}_{\text{pred}}$ is the total time of the predicted task schedule, $\mathcal{T}_{\text{opt}}$ is the total time of the ground-truth optimal schedule obtained by the OR solver, and $\mathcal{T}_{\text{worst}}$ is the total time when all subtasks are executed sequentially without any parallelism. Intuitively, the numerator $\mathcal{T}_{\text{worst}} - \mathcal{T}_{\text{pred}}$ measures the time saved by the model compared to the naive baseline, while the denominator $\mathcal{T}_{\text{worst}} - \mathcal{T}_{\text{opt}}$ is the maximum possible saving for that task instance. Thus, TE reflects the fraction of the theoretically achievable time savings that the model actually realizes, with $\mathrm{TE}=0\%$ indicating purely sequential execution and $\mathrm{TE}=100\%$ matching the optimal schedule.

\begin{table*}[t!]
\centering
\scriptsize
\setlength{\tabcolsep}{1mm}
\begin{minipage}{0.48\textwidth}
\centering
\begin{tabular}{lcccc}
\toprule
 \multirow{2.3}{*}{Setting} & \multicolumn{2}{c}{Language} & \multicolumn{1}{c}{Scheduling}   & \multicolumn{1}{c}{3D Grounding}    \\ 
  \cmidrule(lr){2-3}\cmidrule(lr){4-4}\cmidrule(lr){5-5}
  &METEOR  & ROUGE  & TE  & Accuracy    \\ 
\midrule
No scheduling content      & 35.60  & 48.89 &  21.03    & 15.95\\
+ Scheduling content & 41.29 & 55.28 & 47.04   &  34.74\\ 
+ STM (\textbf{ours}) & \textbf{42.82} & \textbf{62.78}  & \textbf{72.99}    &  \textbf{35.38} \\
\textcolor{gray}{GT scheduling content}  & \textcolor{gray}{53.34} & \textcolor{gray}{75.06} & \textcolor{gray}{90.29}  &  \textcolor{gray}{38.52}  \\ 
\bottomrule
\end{tabular}
\vspace{-5pt}
\caption*{(a) Effect of scheduling token mechanism}
\label{tab:ablation_a}
\end{minipage}
\hfill
\begin{minipage}{0.48\textwidth}
\centering
\setlength{\tabcolsep}{1mm}
\begin{tabular}{l ccc cc}
\toprule
 \multirow{2.3}{*}{Method} & \multicolumn{4}{c}{Subtask number}& \multirow{2.3}{*}{Overall}  \\ 
  \cmidrule(lr){2-5}
  &Four & Five & Six & Seven  \\ 
\midrule
PQ3D~\cite{pq3d} & 14.82 & 14.15 & 13.40 & 13.73 & 14.03\\
LEO~\cite{leo} & 42.14 & 40.12 & 36.42 & 33.91 & 38.14\\
Grounded 3D LLM~\cite{grounded3dllm}  & 54.35 & 45.13 & 36.59 & 36.04 & 43.03\\
\model (\textbf{ours}) &\textbf{ 60.23} & \textbf{52.98} & \textbf{52.03} & \textbf{48.70} & \textbf{53.49}\\
\bottomrule
\end{tabular}
\vspace{-5pt}
\caption*{(b) Performance across task difficulty levels}
\label{tab:ablation_b}
\end{minipage}

\vspace{3pt}

\begin{minipage}{0.48\textwidth}
\centering
\renewcommand{\arraystretch}{0.95}
\setlength{\tabcolsep}{1.6mm}
\begin{tabular}{ccccc}
\toprule
\multirow{2.3}{*}{ \# LLM Params.} & \multicolumn{2}{c}{Language} & \multicolumn{1}{c}{Scheduling}   & \multicolumn{1}{c}{3D Grounding}    \\ 
\cmidrule(lr){2-3}\cmidrule(lr){4-4}\cmidrule(lr){5-5}
  &  METEOR  & ROUGE  & TE  & Accuracy    \\ 
\midrule
1B   & 42.82  & 62.78 &  72.99 & 35.38 \\
7B  & 45.19 & 63.55 & 73.21   &  36.25 \\ 
\bottomrule
\end{tabular}
\vspace{-5pt}
\caption*{(c) Effect of scaling LLM}
\label{tab:ablation_c}
\end{minipage}
\hfill
\begin{minipage}{0.48\textwidth}
\centering
\setlength{\tabcolsep}{2.1mm}
\renewcommand{\arraystretch}{1.5}
\vspace{13pt}
\begin{tabular}{cccccccc}
\toprule
Subtask number & 4 & 5 & 6 & 7 & 10 & 20 & 50 \\
\midrule
Runtime (ms) & 1.14 & 1.28 & 1.31 & 1.42 & 1.49 & 2.01 & 3.94 \\
\bottomrule
\end{tabular}
\vspace{-5pt}
\caption*{(d) Optimization solver runtime}
\label{tab:ablation_d}
\end{minipage}
\vspace{-5pt}
\caption{Ablation studies. (a) Effect of scheduling token mechanism. (b) Performance across different task difficulty levels. (c) Effect of scaling LLM. (d) Runtime of the optimization solver.}
\label{tab:ablation}
\end{table*}

\subsection{Main Results}

We conduct a comprehensive comparison between our proposed \model and existing methods on \dataset dataset, as reported in Tab.~\ref{tab:main}. We evaluate commercial LLM/MLLMs by providing only the text part of the instructions, as they only support images as visual input. Notably, DeepSeek-R1 demonstrates strong performance on task scheduling (TE 72.63\%) due to reinforcement learning on mathematical problems. However, these models cannot process point cloud data or directly locate objects in 3D environments, which limits their applicability in embodied scenarios.

Object-level methods~\cite{3d-vista, pq3d} process on object point clouds obtained from 3D detectors like Mask3D~\cite{mask3d}. While these methods achieve high grounding accuracy, they are limited by their inability to handle long textual inputs and generate multimodal outputs. LEO~\cite{leo} integrates an LLM for enhanced language performance but lacks target object grounding. These methods focus on object-centric 3D understanding, which is insufficient for task-driven embodied scenarios where an agent requires a comprehensive understanding of the entire 3D environment.
For scene-level methods, we use Grounded 3D LLM \cite{grounded3dllm} as the baseline. By introducing the STM, our model achieves a substantial gain (30.53\%) in task scheduling and further boosts 3D grounding by 1.38\%.
Overall, our method consistently outperforms baseline methods, validating its effectiveness across language understanding, 3D grounding, and scheduling efficiency.

We also compare the 3D target object grounding performance of different models in Tab.~\ref{tab:grounding_type}(a). Object-level methods achieve higher performance due to the use of additional 3D object detectors. However, their performance heavily depends on the detector's capability, introducing extra complexity in preprocessing 3D point clouds and leading to the loss of full scene information. In contrast, scene-level methods offer a cleaner alternative for 3D grounding by directly processing the entire scene point cloud, making them more suitable for real-world applications.

Accurate recognition of parallelizable subtasks is a prerequisite for effective time scheduling. As shown in Tab.~\ref{tab:grounding_type}(b), our model achieves the highest accuracy in subtask type recognition, with substantial improvements in recall and F1-score for parallelizable subtasks, which in turn leads to significantly higher time efficiency. Therefore, robust subtask type recognition is critical for enhancing scheduling performance.

\subsection{Ablation Studies} 

\begin{figure}[t]
  \centering
   \includegraphics[width=\linewidth]{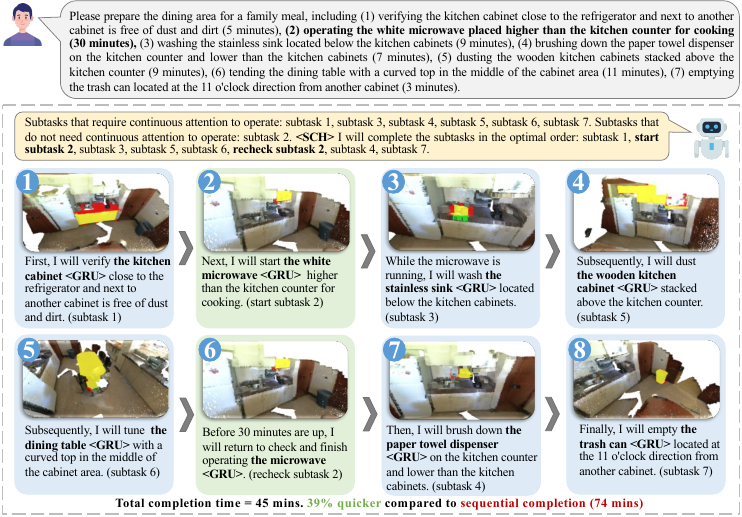}
      \vspace{-15pt}
   \caption{A qualitative example of \model. In the visualized point clouds, \textcolor{DeepYellow}{yellow} shows correct predictions, \textcolor{red}{red} indicates false positives, and \textcolor{deepgreen}{green} marks missed ground truth regions.}
   \label{fig:vis_example}
\end{figure}

\textbf{Effect of STM.}  
As shown in Tab.~\ref{tab:ablation}(a), adding scheduling content substantially improves performance, demonstrating the importance of explicit scheduling. The proposed STM further boosts time efficiency by 25.95\%, highlighting the critical role of proper scheduling. Using ground-truth constraints indicates that more accurate subtask recognition enables additional performance gains.

\textbf{Task difficulty levels.} 
As shown in Tab.~\ref{tab:ablation}(b), performance consistently declines as the number of subtasks increases, indicating that all methods degrade with more complex task structures. This trend highlights the increased challenge in task scheduling and reflects the inherent complexity of the proposed \dataset dataset.

\textbf{Effect of scaling LLM.}
As summarized in Tab.~\ref{tab:ablation}(c), increasing the size of the LLM improves performance across language understanding, task scheduling, and 3D grounding. To balance performance and training cost, we adopt Vicuna-1B as the default setting.

\textbf{Solver runtime.} As the number of subtasks increases, the optimization solver remains extremely fast (Tab.~\ref{tab:ablation}(d)), where even with 50 subtasks the total runtime of the solver stays below 4 ms, introducing virtually no overhead.

\textbf{Qualitative analysis.} Fig.~\ref{fig:vis_example} shows a representative example where our model identifies subtask 2 (microwave operation) as parallelizable and schedules other subtasks during its 30-minute waiting period, saving 29 minutes (39\% efficiency gain) compared to sequential execution. The model also accurately localizes target objects with high IoU, demonstrating effective integration of OR-based scheduling and spatial grounding.

\subsection{Limitation}

While this work demonstrates strong performance on the \dataset benchmark, future work will deploy the framework on physical robots to validate robustness in dynamic environments. Additionally, we will explore integrating the external optimization solver directly within the language model to enable end-to-end differentiable reasoning.

\section{Conclusion}
\label{sec:Conclusion}

In this work, we introduce the \task task that integrates task scheduling from operations research with spatial grounding for embodied agents. We construct the \dataset dataset, and propose \model, an embodied 3D MLLM equipped with the scheduling token mechanism to generate efficient task schedules with grounded actions. Experiments validate the capability of \model across language, grounding, and task scheduling. We believe our initial work on the OR knowledge-intensive scenario can inspire future research to further improve the complex planning capabilities and multi-modal integration of embodied agents.

\section*{Acknowledgments}

This work was supported by the NSFC (Grant No. 62225603 and 623B2038) and in part by the Hubei Provincial Technology Innovation Program (Grant No. 2024BAA007).

\bibliography{aaai2026}

@string{CVPR = "Proceedings of IEEE/CVF Conference on Computer Vision and Pattern Recognition"}

@string{ICCV = "Proceedings of IEEE/CVF International Conference on Computer Vision"}

@string{ECCV = "Proceedings of European Conference on Computer Vision"}

@string{NeurIPS = "Proceedings of the Advances in Neural Information Processing Systems"}

@string{ICML = "Proceedings of the International Conference on Machine Learning"}

@string{AAAI = "Proceedings of the AAAI Conference on Artificial Intelligence"}

@string{ICLR = "Proceedings of the International Conference on Learning Representations"}

@string{TPAMI = "IEEE Transactions on Pattern Analysis and Machine Intelligence"}

@article{sg3d,
  title={Task-oriented sequential grounding in 3d scenes},
  author={Zhang, Zhuofan and Zhu, Ziyu and Li, Pengxiang and Liu, Tengyu and Ma, Xiaojian and Chen, Yixin and Jia, Baoxiong and Huang, Siyuan and Li, Qing},
  journal={arXiv preprint arXiv:2408.04034},
  year={2024}
}

@inproceedings{leo,
  title={An Embodied Generalist Agent in 3D World},
  author={Huang, Jiangyong and Yong, Silong and Ma, Xiaojian and Linghu, Xiongkun and Li, Puhao and Wang, Yan and Li, Qing and Zhu, Song-Chun and Jia, Baoxiong and Huang, Siyuan},
  booktitle=ICML,
  year={2024}
}

@inproceedings{3d-vista,
  title={3D-VisTA: Pre-trained transformer for 3D vision and text alignment},
  author={Zhu, Ziyu and Ma, Xiaojian and Chen, Yixin and Deng, Zhidong and Huang, Siyuan and Li, Qing},
  booktitle=ICCV,
  pages={2911--2921},
  year={2023}
}

@inproceedings{pq3d,
  title={Unifying 3D Vision-Language Understanding via Promptable Queries},
  author={Zhu, Ziyu and Zhang, Zhuofan and Ma, Xiaojian and Niu, Xuesong and Chen, Yixin and Jia, Baoxiong and Deng, Zhidong and Huang, Siyuan and Li, Qing},
  booktitle=ECCV,
  year={2024}
}

@inproceedings{scannet200,
  title={Language-grounded indoor 3d semantic segmentation in the wild},
  author={Rozenberszki, David and Litany, Or and Dai, Angela},
  booktitle=ECCV,
  pages={125--141},
  year={2022}
}

@inproceedings{scanrefer,
  title={Scanrefer: 3d object localization in rgb-d scans using natural language},
  author={Chen, Dave Zhenyu and Chang, Angel X and Nie{\ss}ner, Matthias},
  booktitle=ECCV,
  pages={202--221},
  year={2020}
}

@inproceedings{takmaz2023openmask3d,
  title={{OpenMask3D: Open-Vocabulary 3D Instance Segmentation}},
  author={Takmaz, Ay{\c{c}}a and Fedele, Elisabetta and Sumner, Robert W. and Pollefeys, Marc and Tombari, Federico and Engelmann, Francis},
  booktitle=NeurIPS,
  year={2023}
}

@article{embodied-survey,
  title={A survey of embodied ai: From simulators to research tasks},
  author={Duan, Jiafei and Yu, Samson and Tan, Hui Li and Zhu, Hongyuan and Tan, Cheston},
  journal={IEEE Transactions on Emerging Topics in Computational Intelligence},
  volume={6},
  number={2},
  pages={230--244},
  year={2022},
  publisher={IEEE}
}

@inproceedings{grounded-task,
  title={On grounded planning for embodied tasks with language models},
  author={Lin, Bill Yuchen and Huang, Chengsong and Liu, Qian and Gu, Wenda and Sommerer, Sam and Ren, Xiang},
  booktitle=AAAI,
  volume={37},
  number={11},
  pages={13192--13200},
  year={2023}
}

@article{TaPA,
  title={Embodied task planning with large language models},
  author={Wu, Zhenyu and Wang, Ziwei and Xu, Xiuwei and Lu, Jiwen and Yan, Haibin},
  journal={arXiv preprint arXiv:2307.01848},
  year={2023}
}

@article{LoTa-Bench,
  title={LoTa-Bench: Benchmarking Language-oriented Task Planners for Embodied Agents},
  author={Choi, Jae-Woo and Yoon, Youngwoo and Ong, Hyobin and Kim, Jaehong and Jang, Minsu},
  journal=ICLR,
  year={2024}
}

@article{saycan,
  title={Do as i can, not as i say: Grounding language in robotic affordances},
  author={Ahn, Michael and Brohan, Anthony and Brown, Noah and Chebotar, Yevgen and Cortes, Omar and David, Byron and Finn, Chelsea and Fu, Chuyuan and Gopalakrishnan, Keerthana and Hausman, Karol and others},
  journal={arXiv preprint arXiv:2204.01691},
  year={2022}
}

@article{chat3d,
  title={Chat-3d: Data-efficiently tuning large language model for universal dialogue of 3d scenes},
  author={Wang, Zehan and Huang, Haifeng and Zhao, Yang and Zhang, Ziang and Zhao, Zhou},
  journal={arXiv preprint arXiv:2308.08769},
  year={2023}
}

@article{3dllm,
  title={3d-llm: Injecting the 3d world into large language models},
  author={Hong, Yining and Zhen, Haoyu and Chen, Peihao and Zheng, Shuhong and Du, Yilun and Chen, Zhenfang and Gan, Chuang},
  journal=NeurIPS,
  pages={20482--20494},
  year={2023}
}

@article{hu2021lora,
  title={Lora: Low-rank adaptation of large language models},
  author={Hu, Edward J and Shen, Yelong and Wallis, Phillip and Allen-Zhu, Zeyuan and Li, Yuanzhi and Wang, Shean and Wang, Lu and Chen, Weizhu},
  journal={arXiv preprint arXiv:2106.09685},
  year={2021}
}

@inproceedings{jia2024sceneverse,
  title={SceneVerse: Scaling 3D Vision-Language Learning for Grounded Scene Understanding},
  author={Jia, Baoxiong and Chen, Yixin and Yu, Huangyue and Wang, Yan and Niu, Xuesong and Liu, Tengyu and Li, Qing and Huang, Siyuan},
  booktitle=ECCV,
  year={2024}
}

@article{wang2023embodiedscan,
  title={EmbodiedScan: A Holistic Multi-Modal 3D Perception Suite Towards Embodied AI},
  author={Wang, Tai and Mao, Xiaohan and Zhu, Chenming and Xu, Runsen and Lyu, Ruiyuan and Li, Peisen and Chen, Xiao and Zhang, Wenwei and Chen, Kai and Xue, Tianfan and others},
  journal=CVPR,
  year={2024}
}

@article{baruch2021arkitscenes,
  title={Arkitscenes: A diverse real-world dataset for 3d indoor scene understanding using mobile rgb-d data},
  author={Baruch, Gilad and Chen, Zhuoyuan and Dehghan, Afshin and Dimry, Tal and Feigin, Yuri and Fu, Peter and Gebauer, Thomas and Joffe, Brandon and Kurz, Daniel and Schwartz, Arik and others},
  journal=NeurIPS,
  year={2021}
}

@inproceedings{wald2019rio,
  title={Rio: 3d object instance re-localization in changing indoor environments},
  author={Wald, Johanna and Avetisyan, Armen and Navab, Nassir and Tombari, Federico and Nie{\ss}ner, Matthias},
  booktitle=ICCV,
  pages={7658--7667},
  year={2019}
}

@article{hm3d,
  title={Habitat-matterport 3d dataset (hm3d): 1000 large-scale 3d environments for embodied ai},
  author={Ramakrishnan, Santhosh K and Gokaslan, Aaron and Wijmans, Erik and Maksymets, Oleksandr and Clegg, Alex and Turner, John and Undersander, Eric and Galuba, Wojciech and Westbury, Andrew and Chang, Angel X and others},
  journal=NeurIPS,
  year={2021}
}

@article{mao2022multiscan,
  title={MultiScan: Scalable RGBD scanning for 3D environments with articulated objects},
  author={Mao, Yongsen and Zhang, Yiming and Jiang, Hanxiao and Chang, Angel and Savva, Manolis},
  journal=NeurIPS,
  volume={35},
  pages={9058--9071},
  year={2022}
}

@inproceedings{chatscene,
  title={Chat-scene: Bridging 3d scene and large language models with object identifiers},
  author={Huang, Haifeng and Chen, Yilun and Wang, Zehan and Huang, Rongjie and Xu, Runsen and Wang, Tai and Liu, Luping and Cheng, Xize and Zhao, Yang and Pang, Jiangmiao and others},
  booktitle=NeurIPS,
  year={2024}
}

@article{robin3d,
  title={Robin3D: Improving 3D Large Language Model via Robust Instruction Tuning},
  author={Kang, Weitai and Huang, Haifeng and Shang, Yuzhang and Shah, Mubarak and Yan, Yan},
  journal={arXiv preprint arXiv:2410.00255},
  year={2024}
}

@inproceedings{llava3d,
  title={LLaVA-3D: A Simple yet Effective Pathway to Empowering LMMs with 3D-awareness},
  author={Zhu, Chenming and Wang, Tai and Zhang, Wenwei and Pang, Jiangmiao and Liu, Xihui},
  booktitle=ICCV,
  year={2025}
}

@article{grounded3dllm,
  title={Grounded 3d-llm with referent tokens},
  author={Chen, Yilun and Yang, Shuai and Huang, Haifeng and Wang, Tai and Xu, Runsen and Lyu, Ruiyuan and Lin, Dahua and Pang, Jiangmiao},
  journal={arXiv preprint arXiv:2405.10370},
  year={2024}
}

@article{3dllava,
  title={3D-LLaVA: Towards Generalist 3D LMMs with Omni Superpoint Transformer},
  author={Deng, Jiajun and He, Tianyu and Jiang, Li and Wang, Tianyu and Dayoub, Feras and Reid, Ian},
  journal=CVPR,
  year={2025}
}

@inproceedings{mask3d,
  title={Mask3d: Mask transformer for 3d semantic instance segmentation},
  author={Schult, Jonas and Engelmann, Francis and Hermans, Alexander and Litany, Or and Tang, Siyu and Leibe, Bastian},
  booktitle={2023 IEEE International Conference on Robotics and Automation},
  pages={8216--8223},
  year={2023},
  organization={IEEE}
}

@article{reason3d,
  title={Reason3d: Searching and reasoning 3d segmentation via large language model},
  author={Huang, Kuan-Chih and Li, Xiangtai and Qi, Lu and Yan, Shuicheng and Yang, Ming-Hsuan},
  journal={arXiv preprint arXiv:2405.17427},
  year={2024}
}

@article{uniseg3d,
  title={A unified framework for 3d scene understanding},
  author={Xu, Wei and Shi, Chunsheng and Tu, Sifan and Zhou, Xin and Liang, Dingkang and Bai, Xiang},
  journal={Advances in Neural Information Processing Systems},
  volume={37},
  pages={59468--59490},
  year={2025}
}

@article{unidet3d,
  title={UniDet3D: Multi-dataset Indoor 3D Object Detection},
  author={Maksim Kolodiazhnyi and Anna Vorontsova and Matvey Skripkin and Danila Rukhovich and Anton Konushin},
  journal={arXiv preprint arXiv:2409.04234},
  year={2024}
}

@inproceedings{oneformer3d,
  title={Oneformer3d: One transformer for unified point cloud segmentation},
  author={Kolodiazhnyi, Maxim and Vorontsova, Anna and Konushin, Anton and Rukhovich, Danila},
  booktitle={Proceedings of the IEEE/CVF Conference on Computer Vision and Pattern Recognition},
  pages={20943--20953},
  year={2024}
}

@article{more3d,
  title={Multimodal 3D Reasoning Segmentation with Complex Scenes},
  author={Xueying Jiang and Lewei Lu and Ling Shao and Shijian Lu},
  journal={arXiv preprint arXiv:2411.13927},
  year={2024}
}

@article{integratingchainofthought,
  title={Integrating Chain-of-Thought for Multimodal Alignment: A Study on 3D Vision-Language Learning},
  author={Chen, Yanjun and Sun, Yirong and Chen, Xinghao and Wang, Jian and Shen, Xiaoyu and Li, Wenjie and Zhang, Wei},
  journal={arXiv preprint arXiv:2503.06232},
  year={2025}
}

@inproceedings{palme,
    title={PaLM-E: An Embodied Multimodal Language Model},
    author={Driess, Danny and Xia, Fei and Sajjadi, Mehdi S. M. and Lynch, Corey and Chowdhery, Aakanksha and Ichter, Brian and Wahid, Ayzaan and Tompson, Jonathan and Vuong, Quan and Yu, Tianhe and Huang, Wenlong and Chebotar, Yevgen and Sermanet, Pierre and Duckworth, Daniel and Levine, Sergey and Vanhoucke, Vincent and Hausman, Karol and Toussaint, Marc and Greff, Klaus and Zeng, Andy and Mordatch, Igor and Florence, Pete},
    booktitle=ICML,
    year={2023}
}

@article{deepseekr1,
  publtype={informal},
  author={Daya Guo and Dejian Yang and Haowei Zhang and Junxiao Song and Ruoyu Zhang and Runxin Xu},
  title={DeepSeek-R1: Incentivizing Reasoning Capability in LLMs via Reinforcement Learning},
  year={2025},
  month={January},
  cdate={1735689600000},
  journal={Nature}
}

@inproceedings{scanreason,
  title={Scanreason: Empowering 3d visual grounding with reasoning capabilities},
  author={Zhu, Chenming and Wang, Tai and Zhang, Wenwei and Chen, Kai and Liu, Xihui},
  booktitle=ECCV,
  pages={151--168},
  year={2024}
}

@inproceedings{
intent3d,
title={Intent3D: 3D Object Detection in {RGB}-D Scans Based on Human Intention},
author={Weitai Kang and Mengxue Qu and Jyoti Kini and Yunchao Wei and Mubarak Shah and Yan Yan},
booktitle=ICLR,
year={2025}
}

@misc{vicuna,
    title = {Vicuna: An Open-Source Chatbot Impressing GPT-4 with 90\%* ChatGPT Quality},
    url = {https://lmsys.org/blog/2023-03-30-vicuna/},
    author = {Chiang, Wei-Lin and Li, Zhuohan and Lin, Zi and Sheng, Ying and Wu, Zhanghao and Zhang, Hao and Zheng, Lianmin and Zhuang, Siyuan and Zhuang, Yonghao and Gonzalez, Joseph E. and Stoica, Ion and Xing, Eric P.},
    month = {March},
    year = {2023}
}

@article{EgoPlan-Bench,
  title={Egoplan-bench: Benchmarking multimodal large language models for human-level planning},
  author={Chen, Yi and Ge, Yuying and Ge, Yixiao and Ding, Mingyu and Li, Bohao and Wang, Rui and Xu, Ruifeng and Shan, Ying and Liu, Xihui},
  journal={arXiv preprint arXiv:2312.06722},
  year={2023}
}

@article{xu2025igev++,
  title={Igev++: Iterative multi-range geometry encoding volumes for stereo matching},
  author={Xu, Gangwei and Wang, Xianqi and Zhang, Zhaoxing and Cheng, Junda and Liao, Chunyuan and Yang, Xin},
  journal=TPAMI,
  year={2025},
  publisher={IEEE}
}

@article{xu2023accurate,
  title={Accurate and efficient stereo matching via attention concatenation volume},
  author={Xu, Gangwei and Wang, Yun and Cheng, Junda and Tang, Jinhui and Yang, Xin},
  journal=TPAMI,
  volume={46},
  number={4},
  pages={2461--2474},
  year={2023},
  publisher={IEEE}
}

@inproceedings{zhou2025hermes,
  title={Hermes: A unified self-driving world model for simultaneous 3d scene understanding and generation},
  author={Zhou, Xin and Liang, Dingkang and Tu, Sifan and Chen, Xiwu and Ding, Yikang and Zhang, Dingyuan and Tan, Feiyang and Zhao, Hengshuang and Bai, Xiang},
  booktitle=ICCV,
  year={2025}
}

@article{liang2025parameter,
  title={Parameter-efficient fine-tuning in spectral domain for point cloud learning},
  author={Liang, Dingkang and Feng, Tianrui and Zhou, Xin and Zhang, Yumeng and Zou, Zhikang and Bai, Xiang},
  journal=TPAMI,
  year={2025},
  publisher={IEEE}
}

@article{liang2025sood++,
  title={Sood++: Leveraging unlabeled data to boost oriented object detection},
  author={Liang, Dingkang and Hua, Wei and Shi, Chunsheng and Zou, Zhikang and Ye, Xiaoqing and Bai, Xiang},
  journal=TPAMI,
  year={2025},
  publisher={IEEE}
}

@inproceedings{liang2024pointmamba,
  title={Pointmamba: A simple state space model for point cloud analysis},
  author={Liang, Dingkang and Zhou, Xin and Xu, Wei and Zhu, Xingkui and Zou, Zhikang and Ye, Xiaoqing and Tan, Xiao and Bai, Xiang},
  booktitle=NeurIPS,
  volume={37},
  pages={32653--32677},
  year={2024}
}

@inproceedings{fu2025orion,
  title={Orion: A holistic end-to-end autonomous driving framework by vision-language instructed action generation},
  author={Fu, Haoyu and Zhang, Diankun and Zhao, Zongchuang and Cui, Jianfeng and Liang, Dingkang and Zhang, Chong and Zhang, Dingyuan and Xie, Hongwei and Wang, Bing and Bai, Xiang},
  booktitle=ICCV,
  year={2025}
}

\end{document}